\documentclass{styles/svproc}
\usepackage[linkcolor=black,citecolor=black,urlcolor=black,colorlinks=true]{hyperref}
\usepackage{url}
\usepackage[space,sort]{cite}

\usepackage{graphicx}
\usepackage{subfigure}
\usepackage[font=small,labelfont=bf]{caption}
\usepackage{epstopdf}
\usepackage{pdflscape}
\usepackage{amsmath,amssymb,bm}
\usepackage{algorithmic}
\usepackage{comment} 
\usepackage[dvipsnames]{xcolor}
\begin{document}
\mainmatter              
\title{\textit{\small Accepted to Distributed Autonomous Robotic Systems 2024 (DARS) 2024}\\[1em]
Multi-Robot Target Tracking with Sensing and Communication Danger Zones}
\titlerunning{Target Tracking with Danger Zones}  
%
\author{Jiazhen Liu\inst{1}$^\star$ \and Peihan Li\inst{2}$^\star$\and
Yuwei Wu\inst{3} \and  \\ Gaurav S. Sukhatme\inst{4} \and Vijay Kumar\inst{3} \and Lifeng Zhou\inst{2}$^\dagger$ \\ $^\star$ Equally contributed \\$^\dagger$ Corresponding author}
\authorrunning{Jiazhen Liu et al.} 
%
\tocauthor{Jiazhen Liu, Peihan Li, Yuwei Wu, Gaurav S. Sukhatme, Vijay Kumar, Lifeng Zhou
}
%

\institute{Georgia Institute of Technology, Georgia GA 30332, USA, \email{jliu3103@gatech.edu}
\and
Drexel University, Philadelphia PA 19104, USA, 
\email{\{pl525,lz457\}@drexel.edu}
\and 
University of Pennsylvania, Philadelphia PA 19104, USA,
\email{\texttt{\small \{yuweiwu, kumar\}@seas.upenn.edu}}
\and 
University of Southern California,  Los Angeles CA 90089, USA, 
\email{gaurav@usc.edu}.
}


\maketitle      
\begin{abstract}
Multi-robot target tracking finds extensive applications in different scenarios, such as environmental surveillance and wildfire management, which require the robustness of the practical deployment of multi-robot systems in uncertain and dangerous environments.
Traditional approaches often focus on the performance of tracking accuracy with no modeling and assumption of the environments, 
neglecting potential environmental hazards which result in system failures in real-world deployments.
To address this challenge, we investigate multi-robot target tracking in the adversarial environment considering sensing and communication attacks with uncertainty.  
We design specific strategies to avoid different danger zones and propose a multi-agent tracking framework under the perilous environment.
We approximate the probabilistic constraints and formulate practical optimization strategies to address computational challenges efficiently. 
We evaluate the performance of our proposed methods in simulations to demonstrate the ability of robots to adjust their risk-aware behaviors under different levels of environmental uncertainty and risk confidence.
The proposed method is further validated via real-world robot experiments where a team of drones successfully track dynamic ground robots while being risk-aware of the sensing and/or communication danger zones.

\end{abstract}

\vspace{-0.5cm}
\section{Introduction}





Multi-robot active target tracking refers to the problem of planning the (joint) motion of a team of robots to optimize certain tracking objectives. It has wide applications in surveillance~\cite{rao1993fully}, environmental monitoring~\cite{dunbabin2012robots}, and wildfire covering~\cite{8206579}, and a wealth of approaches have been proposed to tackle this problem~\cite{7139863,spletzer2003dynamic,tokekar2014multi,dames2017detecting,10018712}. These approaches can generally be categorized into centralized~\cite{spletzer2003dynamic,dames2017detecting} and decentralized ones~\cite{7139863,10018712}. Submodular functions (i.e., functions that have a diminishing return property) are often used as the tracking objective. To leverage the properties of submodularity, greedy algorithms with suboptimality guarantees have been derived~\cite{10018712,7139863}. 

Deployment of multiple robots to track real targets in practical scenarios remains challenging, as the environment or targets themselves can be dangerous or adversarial. Robots may be subject to attacks resulting in sensor damage, communication interruption, or other system faults. Therefore, tracking performance is no longer the only factor to consider, as the robots must also pay special attention to secure themselves from attacks.~\cite{8534468,9763059,liu2022decentralized,8593630,9726858} have closely investigated multi-robot target tracking in the adversarial setting.~\cite{9763059} designed a distributed strategy based on divide-and-conquer that can protect robots from denial-of-service (DoS) attacks. However, it considers attacks in the worst case and overestimates the number of attacks.~\cite{9763059} then improved the overly conservative strategy with an algorithm that can infer the number of DoS attacks using information from 3-hop neighbors.~\cite{8534468} similarly focused on worst-case attacks but showed that the algorithm still gives superior performance even when robots are under non-worst-case attacks (such as random attacks).

Though awareness of risk has been investigated in the context of multi-robot systems in these prior works, the worst-case assumption is often adopted, leading to excessively conservative decision-making. In this paper, we alternatively consider risk in a probabilistic sense and model the safety requirement for robots, i.e., the risk level must not exceed a threshold, using chance-based constraints. Our problem is relevant to motion/trajectory planning under uncertainty~\cite{5970128,8613928,gopalakrishnan2017chance,8202163,8315107}.~\cite{5970128} models an autonomous vehicle as a linear Gaussian system and imposes safety constraints for it by formulating a disjunctive convex optimization. 
\cite{8613928} presented a probabilistic collision avoidance method to navigate micro aerial vehicles through dynamic obstacles. Linear approximation of the probabilistic collision constraints is applied such that the algorithm runs in real time. Inspired by these works, we apply the approximation techniques to alleviate the computational hardness of evaluating the risk constraints.
Our contributions can be summarized as : 
\vspace{-0.85cm}
\begin{itemize}
    \item We propose a novel multi-robot active target tracking framework in environments characterized by \textbf{danger zones}. We categorize adversarial attacks into two primary types: one that induces failures of robots' equipped sensors, and one that jams the communication channels between robots. 
    \item We formulate the tracking problem with danger zone conditions into a nonlinear optimization problem.
    We design different safe distance conditions towards two types of danger zones as probabilistic constraints and provide practical approximations for efficient online planning. 
    \item We conduct thorough evaluations of our framework in simulations. Robots show risk-aware behaviors in the face of the sensing and/or communication danger zones with various uncertainty levels and risk requirements. Additionally, we validate the robustness and effectiveness of our approach in hardware experiments utilizing a team of Crazyflie drones to track multiple ground robots.
\end{itemize}

\section{Problem Formulation}

\vspace{-0.2cm}
\subsection{Assumptions and Notations}

We consider the problem of active target tracking with $M$ robots and $N$ targets. The goal is to enable the robots to minimize their estimation uncertainty of the target positions while simultaneously securing themselves from attacks. Both planning and target position estimation are carried out in a centralized manner.
We consider two different types of danger sources: sensing attack source and communication attack source. Both types are static sources. We introduce the details in this and the next subsection. 
We use $[n]$ to denote the set of consecutive integers. 

\vspace{-0.4cm}
\subsection{Sensing danger zone}
\label{sensing danger zone}

Let us assume there are $p$ sensing attack sources in the environment.
The sensing attack source initiates attacks on robot perception infrastructure, such as cameras and LiDAR, if the robot is within a known safety clearance $r_l$ to the source $l$,  $l \in [p]$. 
Such a disk-shaped region is defined as a sensing danger zone. 
Thus, the collection of sensing danger zones is denoted by $\mathcal{S}_1, \mathcal{S}_2, \cdots, \mathcal{S}_p$.

The positions of danger sources may come from imprecise estimates, such as those obtained from top-view satellite images. In practical scenarios where the complete environment state is inaccessible, the locations of sensing attack sources are considered only partially revealed to the multi-robot team.
To address this, we characterize the position of the $l^{th}$ sensing attack source $\mathbf{x}_{\mathcal{S}_l} \in \mathbb{R}^2$ with a Gaussian distribution $\mathcal{N}(\mathbf{\mu}_{\mathcal{S}_l}, \mathbf{\Sigma}_{\mathcal{S}_l})$. Information of $\mathbf{\mu}_{\mathcal{S}_l}, \mathbf{\Sigma}_{\mathcal{S}_l}$ is known, but exact value of  $\mathbf{x}_{\mathcal{S}_l}$ remains unknown.
Given that moving closer to the danger source puts the sensor suite under threat, the robots are discouraged from being too close to the danger sources. 
Under the uncertain setting, this requirement is naturally modeled as a chance constraint. For each robot $i\in[M]$ with position $\mathbf{x}_i$, the constraint is, 
\begin{equation}
    \text{Prob}(\mathbf{x}_i \in \mathcal{S}_l) \leq \epsilon_1, \forall i, \forall l,
    \label{eq:prob_dist_small}
\end{equation}
which bounds the probability of entering any sensing danger zone $\mathcal{S}_l$. $\epsilon_1 \in (0, 1)$ is the confidence level. Equivalently, a robot is within the danger zone if the distance between it and the source is less than the safety clearance, i.e., $\lVert \mathbf{x}_{\mathcal{S}_l} - \mathbf{x}_i \rVert \leq r_l$. Hence, the probability above can be computed as
\begin{equation}
\begin{aligned}
    \text{Prob}(\mathbf{x}_i \in \mathcal{S}_l) &= \int_{\lVert \mathbf{x}_{\mathcal{S}_l} - \mathbf{x}_i \rVert \leq r_l} \text{pdf}(\mathbf{x}_{\mathcal{S}_l} - \mathbf{x}_i)d(\mathbf{x}_{\mathcal{S}_l} - \mathbf{x}_i) \\
    &= \int_{\lVert \mathbf{w}_{i,\mathcal{S}_l} \rVert \leq r_l} \text{pdf}(\mathbf{w}_{i,\mathcal{S}_l})d\mathbf{w}_{i,\mathcal{S}_l},\label{eq:integral_prob_typeI}
\end{aligned}
\end{equation}
where $\text{pdf}$ is the probability density function. We have $\mathbf{w}_{i, \mathcal{S}_l} = \mathbf{x}_{\mathcal{S}_l} - \mathbf{x}_i$, and $\mathbf{w}_{i, \mathcal{S}_l}$ also follows a Gaussian distribution $\mathcal{N}(\mathbf{\mu}_{\mathcal{S}_l}~-~\mathbf{x}_i,~\mathbf{\Sigma}_{\mathcal{S}_l})$. 
The probability is an integral of a multivariate Gaussian random variable over a disk-shaped region  $\{\mathbf{w}|\lVert \mathbf{w}_{i,\mathcal{S}_l} \rVert \leq r_l \}$, which lacks a closed form solution and thus cannot be directly computed. 

\vspace{-0.4cm}
\subsection{Communication danger zone}
\label{communication danger zone}
Communication danger zones are areas where \textit{jamming} happens. The jamming effect refers to the situation where an antagonistic agent can send deceiving or noisy signals to interfere with the robot's normal communication with the central station or inter-robot message passing. 
Such interference can potentially deny robots from crucial services such as localization. 
We define communication danger zones as disk-shaped regions, each centered around a jamming source. 
The exact positions of the jamming sources are unknown. We present a formulation of the constraint for reducing the jamming effect precautiously. Note that the focus of this paper is not resilient behavior once communication attacks have truly happened. 

There are $q$ communication danger zones in the environment, $\mathcal{C}_1, \mathcal{C}_2, \cdots, \mathcal{C}_q$. $\mathbf{x}_{\mathcal{C}_k} \in \mathbb{R}^2$ is the position of a jamming source in the $k^{\text{th}}$ zone for $k \in [q]$. $\mathbf{x}_{\mathcal{C}_k}$ is uncertain and modeled as a multivariate Gaussian random variable, $\mathbf{x}_{\mathcal{C}_k}\sim \mathcal{N}(\mathbf{\mu}_{\mathcal{C}_k}, \mathbf{\Sigma}_{\mathcal{C}_k})$, where $\mathbf{\mu}_{\mathcal{C}_k}$ and $\mathbf{\Sigma}_{\mathcal{C}_k}$ are known to the multi-robot team. 

\vspace{-0.5 cm}
\begin{figure}[!h]
    \centering
    \captionsetup[subfigure]{font=scriptsize, labelfont=scriptsize}
    \subfigure[]{
      \includegraphics[width=0.17\columnwidth]{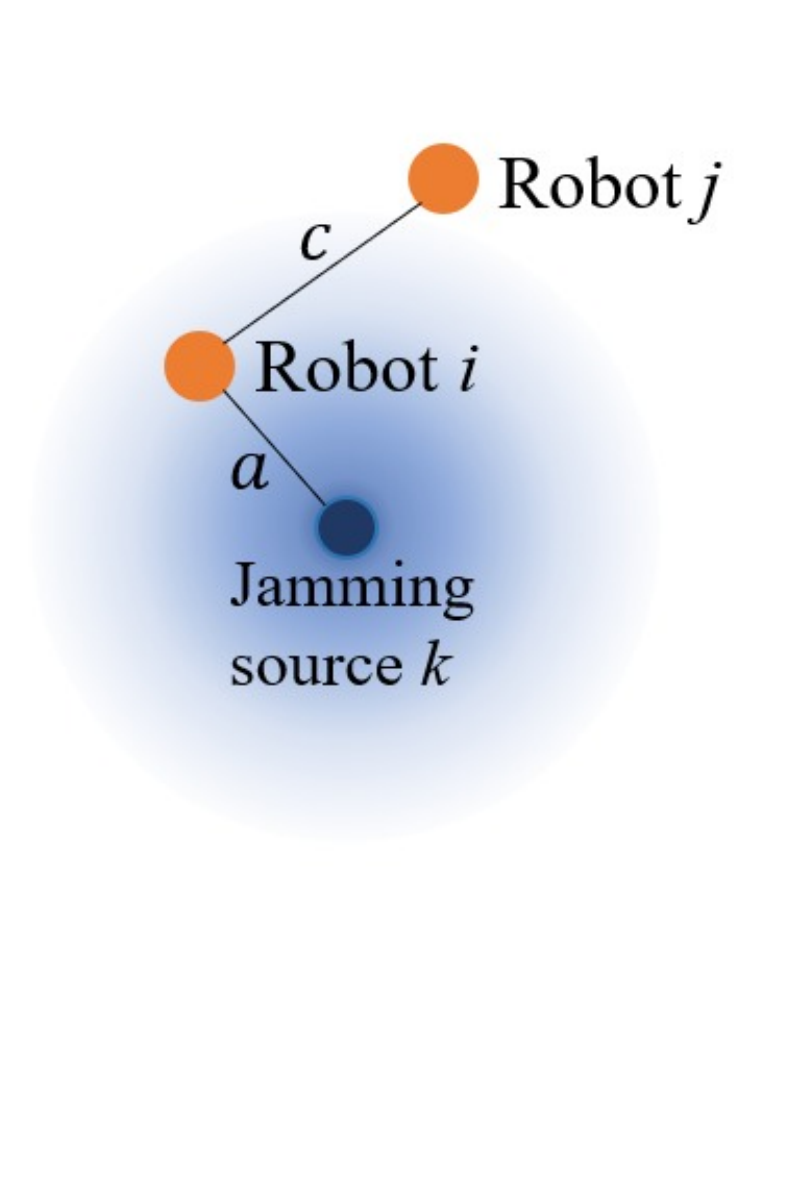}
      \label{fig:geometric_relation}
    }
    \subfigure[]{
      \includegraphics[width=0.25\columnwidth]{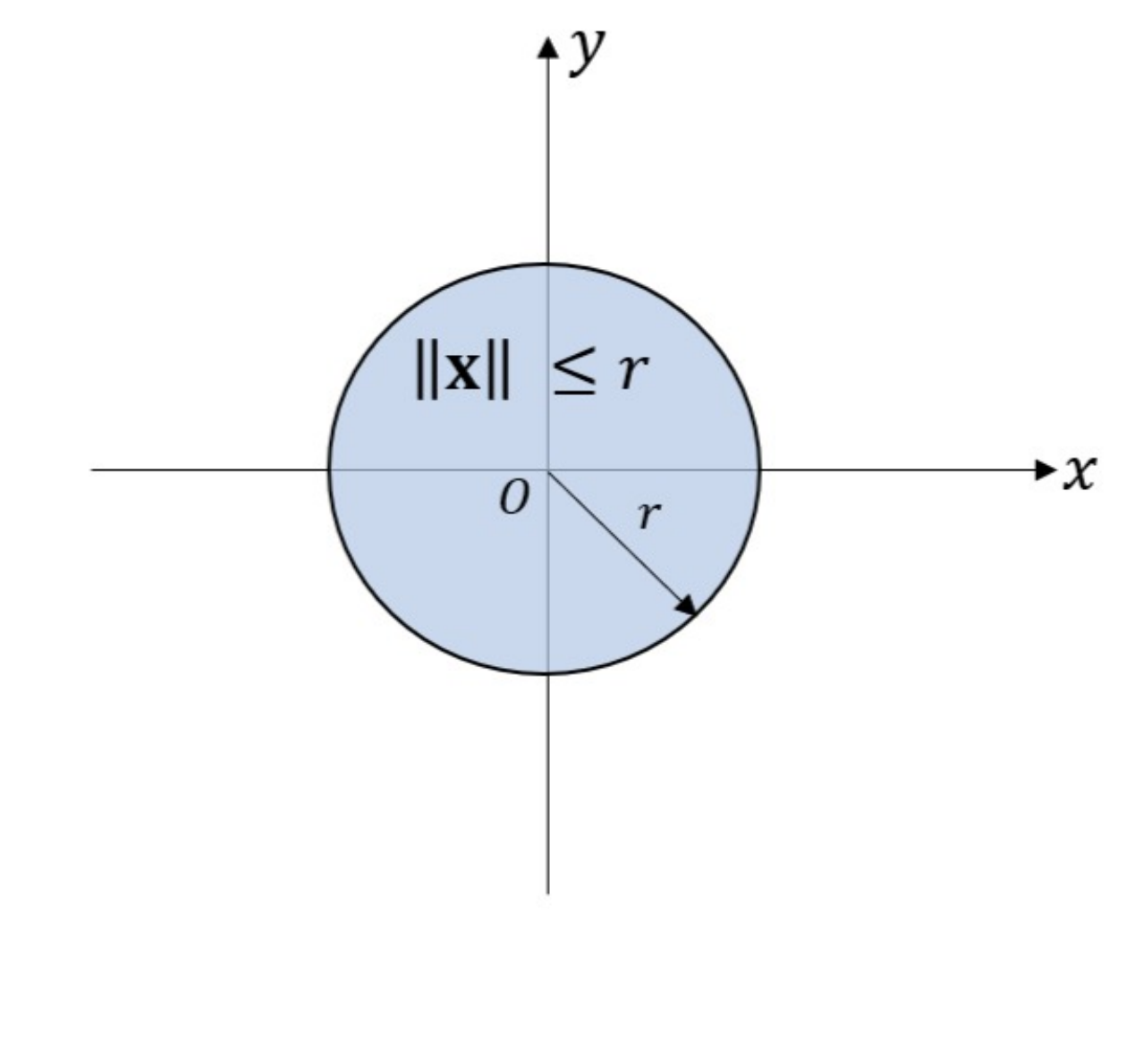}
      \label{fig:integrate_circle}
    }
    \subfigure[]{
      \includegraphics[width=0.25\columnwidth]{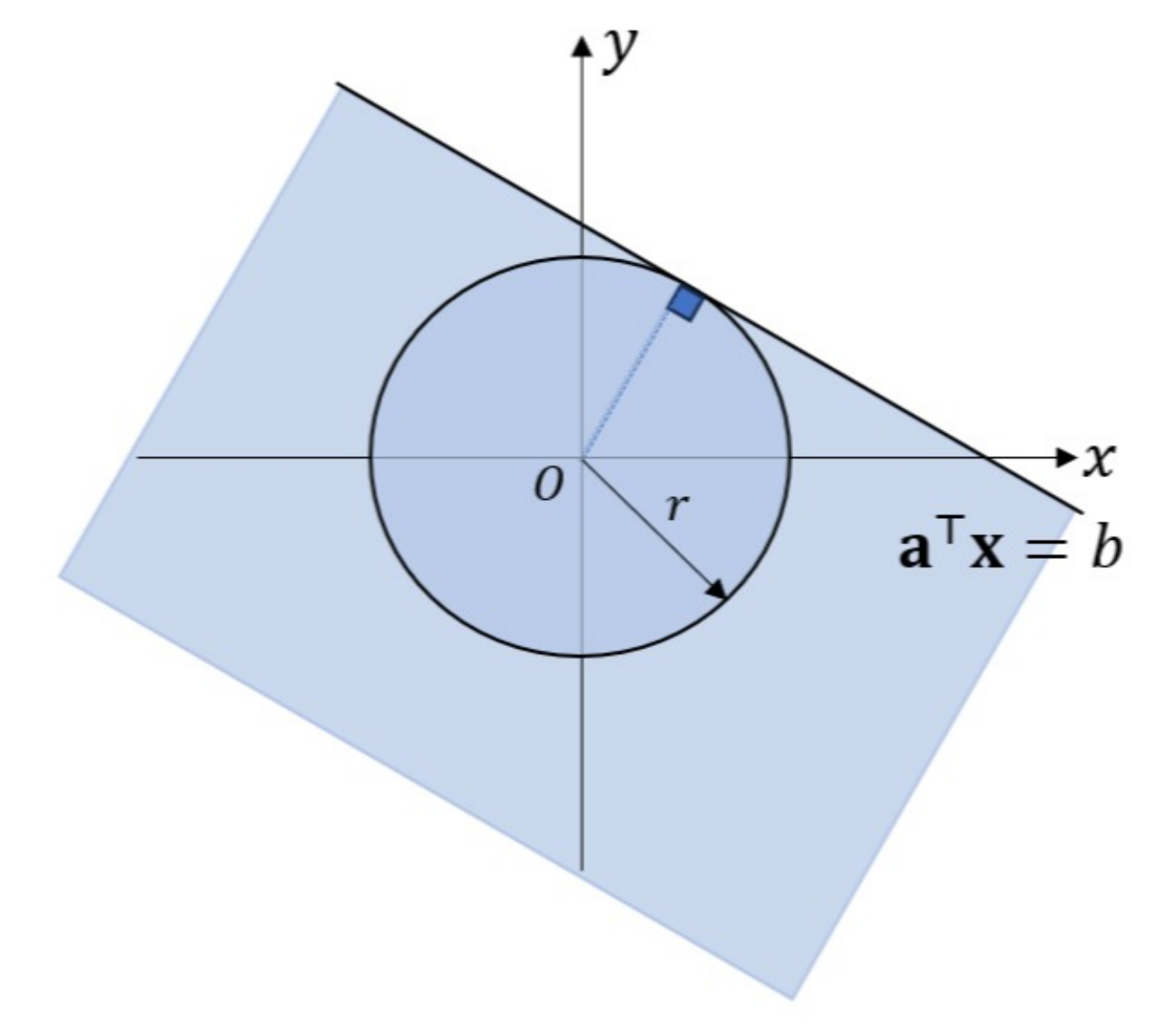}
      \label{fig:integrate_approx}
    }
    \caption{(a) Geometrical illustration for a communication danger zone where the position of the central attacking source is non-deterministic. The uncertainty is reflected as a blue radial gradient; (b) Shows the original integration region corresponding to the chance constraints; (c) Linearizes the integration region to compute an upper-bound of the multivariate integral. The line $\mathbf{a}^\top \mathbf{x} = b$ is tangent to the circle. }
 \label{fig:approximation}
 \vspace{-0.6 cm}
\end{figure}

Next, we show the \textit{geometric conditions} under which the jamming attack will be successful, given that the exact positions of the jamming sources are non-deterministic.
We use robot $i$, its teammate $j$ for $j \neq i$, and the jamming source in zone $\mathcal{C}_k$ as an example, but the idea applies to all triplets. As shown in Fig.~\ref{fig:geometric_relation}, when robot $i$ is inside a communication danger zone $\mathcal{C}_k$, the jamming source $k$ can attack the communication channel between robot $i$ and its teammate $j$. 
Let the distance between robot $i$ and $j$ be $c$, and the distance between robot $i$ and the jamming source $k$ be $a$. 
We encode the degree to which robot $i$ and $j$ can maintain reliable communication using the ratio of distances~\cite{safak2017digital,1637931}, defined as:
\begin{equation}
    \gamma_{ijk} = \frac{a}{c}.
    \label{eq:sj_ratio}
\end{equation}
With an adjustable parameter $\gamma_{ijk}$, we require $\gamma_{ijk} \geq \delta_2$ to increase the chance that robot $i$ reliably maintains communication with robot $j$, since it is desirable for robot $i$ to be close to $j$ and away from the jamming source. Therefore, we impose a lower bound on the ratio between robot $i$'s distance to the jamming source $k$ and its distance to robot $j$. 
Considering that $\mathbf{x}_{\mathcal{C}_k}$ is a random variable, we use a chance constraint to enforce the lower bound formally: 
\begin{equation}
\begin{aligned}
    & \text{Prob}(\frac{a}{c} \geq \delta_2) \geq 1-\epsilon_2,  \iff &  \text{Prob}(\frac{a}{c} < \delta_2) \leq \epsilon_2,
    \label{eq:comm_constraint_chance}
\end{aligned}
\end{equation}
where $\epsilon_2 \in (0, 1)$ is the risk threshold (confidence level) to ensure that the probability of communication between robot $i$ and robot $j$ being jammed is lower than $\epsilon_2$.

Let $\mathbf{x}_i$, $\mathbf{x}_j$, and $\mathbf{x}_{\mathcal{C}_k}$ denote the positions of robot $i$, robot $j$, and the jamming source in $\mathcal{C}_k$, respectively. Since $\mathbf{x}_{\mathcal{C}_k}$ follows a Gaussian distribution, we know that $\mathbf{v}_{i, \mathcal{C}_k} = \mathbf{x}_{\mathcal{C}_k} - \mathbf{x}_i$ also follows a Gaussian distribution $\mathcal{N}(\mathbf{\mu}_{\mathcal{C}_k} - \mathbf{x}_i, \mathbf{\Sigma}_{\mathcal{C}_k})$. 
The probability in Eq.~\ref{eq:comm_constraint_chance} is computed as:
\begin{equation}
\begin{aligned}
    \text{Prob}(\frac{a}{c} < \delta_2) & =  \text{Prob}(a < \delta_2 c), \\ & = \int_{\lVert \mathbf{x}_{\mathcal{C}_k} - \mathbf{x}_i \rVert < \delta_2 c}\text{pdf}(\mathbf{x}_{\mathcal{C}_k} - \mathbf{x}_i )d(\mathbf{x}_{\mathcal{C}_k} - \mathbf{x}_i) ,
    \\ & = \int_{\lVert \mathbf{v}_{i, \mathcal{C}_k}\rVert < \delta_2 c}\text{pdf}(\mathbf{v}_{i, \mathcal{C}_k})d\mathbf{v}_{i, \mathcal{C}_k},
    \label{eq:comm_prob_integrate} 
\end{aligned}
\end{equation}
Eq.~\ref{eq:comm_prob_integrate} integrates the probability density function of a multivariate Gaussian variable $\mathbf{v}_{i, \mathcal{E}_k}$ across a disk that is centered at the origin and has radius $\delta_2 c$. 
Each teammate of the robot $i$ will form a corresponding constraint. We define $c^*$ as
\begin{equation}
    c^* = \text{max}\{ c_1, c_2, \cdots, c_M\}, \label{eq:max_dist}
\end{equation}
which corresponds to the largest disk across which we perform the integration in Eq.~\ref{eq:comm_prob_integrate} for robot $i$. To protect robot $i$ from jamming attack in danger zone $\mathcal{C}_k$, we constraint the upper bound with the threshold, as
\begin{equation}
    \text{Prob}(a < \delta_2 c) \leq \text{Prob}(a < \delta_2 c^*)\leq \epsilon_2.\label{eq:comm_upb}
\end{equation}

\subsection{Target Tracking Optimization with Danger Zones}

With two types of danger zones introduced, we now present a nonlinear optimization program in Eq.~\ref{eq:opt} to model \textit{online} target tracking under the threats of sensing and communication attacks. Eq.~\ref{eq:opt} optimizes tracking performance in the objective function and incorporates safety constraints originating from the danger zones. The optimization over control inputs $\mathbf{u}_{t} = [\mathbf{u}_{1,t}, \dots, \mathbf{u}_{M,t}]^{\intercal}$ for the whole robot team is solved as follows, 
\begin{subequations}
    \label{eq:opt}
    \begin{align}
        & \min_{\mathbf{u}_{t}} \ \    w_1  \cdot f(\mathbf{x}_{t+1}, \hat{\mathbf{z}}_{t+1}) + w_2\cdot \sum_{i=1}^{M}\lVert \mathbf{u}_{i,t}\rVert \quad \ \ \ \label{eq:opt_objective}\\ 
        \textrm{s.t.} \
        & \mathbf{x}_{t+1} = \mathbf{\Phi}\mathbf{x}_{t} + \mathbf{\Lambda}\mathbf{u}_{t},  \quad \label{eq:opt_dynamics}\\
        &\text{Prob}(\lVert \mathbf{x}_{\mathcal{S}_l} - \mathbf{x}_{i, t+1} \rVert \leq r_l) \leq \epsilon_1, \forall i \in [M], \ \forall l \in [p], \ \ \label{eq:opt_constraint_1}\\
         & \text{Prob}(a_{ik} < \delta_2 c_{i}^*)\leq \epsilon_2, \forall i \in [M], \ \forall k \in [q]. \  
 \label{eq:opt_constraint_2}
        \end{align}
\end{subequations}
The objective function balances between minimizing tracking error (first term) and control efforts (second term) with constant weights $w_1, w_2$. $f(\cdot,\cdot)$ represents the uncertainty in target state estimation. It is a nonlinear function dependent on robot positions at time $t+1$ and predicted target positions at time $t+1$, i.e. $\mathbf{x}_{t+1}$ and $\hat{\mathbf{z}}_{t+1}$. Eq.~\ref{eq:opt_dynamics}~describes the dynamics constraint of all the robots. $\mathbf{x}_{t+1}=[\mathbf{x}_{1,t+1}, \dots, \mathbf{x}_{M,t+1}]^{\intercal}$ is the state of all robots at step $t+1$. We adopt a simple linear dynamics model, with process matrix $\mathbf{\Phi}$ and control matrix $\mathbf{\Lambda}$, but switching to more complex ones is straightforward. Eq.~\ref{eq:opt_constraint_1} is the constraint preventing robots from entering sensing danger zones. Eq.~\ref{eq:opt_constraint_2} is the constraint to preserve effective inter-robot communication in communication danger zones. We use $a_{ik}$ to denote the distance between robot $i$ and the jamming source of communication danger zone $\mathcal{C}_k$, and $c_i^*$ to denote the $c^*$ value of robot $i$ (see Eq.~\ref{eq:max_dist}), both calculated using robot $i$'s position at time step $t+1$, i.e., $\mathbf{x}_{i, t+1}$. 
This problem is challenging to solve directly with probabilistic constraints (\ref{eq:opt_constraint_1})(\ref{eq:opt_constraint_2}) and the objective function involved the evaluation towards robots' position $\mathbf{x}_{t+1}$ and predicted positions of targets $\hat{\mathbf{z}}_{t+1}$. 
In the following section, we will discuss the reformulation and approximation of the optimization problem.

\vspace{-0.4cm}
\section{Approach}\label{sec:approach}



\subsection{Chanced-based Constraints Approximation}
\label{sec:chance_constraints_approx}
To address the computational challenge posed by integrating Eq.~\ref{eq:opt_constraint_1} and Eq.~\ref{eq:opt_constraint_2}, we utilize an upper-bound approximation for the probabilities. 
This enables us to convert the chance-based constraints into deterministic constraints.

We begin by revisiting the approximation of the constraints in~\cite{5970128,8613928}. 
Let $\mathbf{x} \in \mathbb{R}^{n_x}$ represent a random variable following a Gaussian distribution, $\mathbf{x} \sim \mathcal{N}(\mathbf{\mu}, \mathbf{\Sigma})$. 
We consider probablistic constraint of the form $\text{Prob}(\mathbf{a}^\top\mathbf{x} \leq b) \leq \delta$, where $\mathbf{a} \in \mathbb{R}^{n_x}$, $b\in\mathbb{R}$ are constants, and $\delta \in (0, 0.5)$ denotes  the confidence level.  
The probabilistic constraint can be transformed into a deterministic one: 
\begin{equation}
    \text{Prob}(\mathbf{a}^\top \mathbf{x} \leq b) \leq \delta \iff \mathbf{a}^\top \mathbf{\mu} - b \geq \gamma,
\label{eq: chance}
\end{equation}
where $\gamma = \text{erf}^{-1}(1-2\delta)\sqrt{2\mathbf{a}^\top\mathbf{\Sigma}\mathbf{a}}$, $\text{erf}(\cdot)$ is the standard error function. 

The chance constraints in Eq.~\ref{eq:opt_constraint_1} and Eq.~\ref{eq:opt_constraint_2} can be converted into deterministic constraints which consequently simplify the computation process. 
The integration areas for both types of danger zones are disk-shaped, as illustrated in Fig.~\ref{fig:integrate_circle}.
We employ a linear approximation of the integration area to calculate a conservative upper bound of the probability. 
This approximation is demonstrated in Fig.~\ref{fig:integrate_approx}, where the integration boundary is approximated by a line $\mathbf{a}^\top \mathbf{x} = b$ tangent to the disk.


To ensure safety in a sensing danger zone with its attacking source at $\mathbf{x}_{\mathcal{S}_l}\sim\mathcal{N}(\mathbf{\mu}_{\mathcal{S}_l}, \mathbf{\Sigma}_{\mathcal{S}_l})$, we linearize the constraint as

\begin{equation}
    \text{Prob}(\mathbf{x}_i \in \mathcal{S}_l) \leq \text{Prob}(\mathbf{a}_{i,\mathcal{S}_l}^\top\mathbf{w}_{i, \mathcal{S}_l} \leq r_l) \leq \epsilon_1.\label{eq:prob_ub_typeI}
\end{equation}
In other words, it requires the approximate probability to be bounded by $\epsilon_1$. $\mathbf{a}_{i, \mathcal{S}_l} = (\mathbf{\mu}_{\mathcal{S}_l}-\mathbf{x}_i) / \lVert \mathbf{\mu}_{\mathcal{S}_l}-\mathbf{x}_i \rVert$ is a unit vector. 
The chance constraint in Eq.~\ref{eq:prob_ub_typeI} has the same form as Eq.~\ref{eq: chance}, and thus is equivalent to a deterministic constraint,
\begin{equation}
    \mathbf{a}_{i, \mathcal{S}_l}^\top (\mathbf{\mu}_{\mathcal{S}_l} - \mathbf{x}_i) - r_l \geq \text{erf}^{-1}(1-2\epsilon_1)\sqrt{2\mathbf{a}_{i, \mathcal{S}_l}^\top\mathbf{\Sigma}_{\mathcal{S}_l}\mathbf{a}_{i, \mathcal{S}_l}}.\label{eq:type_I_determin}
\end{equation}
For simplicity, we express this constraint as $g_{\mathcal{S}_l}(\mathbf{x}_i) \geq 0$. 
We apply the same approximation approach to constraints for communication danger zones, 
so that the probability in Eq.~\ref{eq:opt_constraint_2} becomes
\begin{equation}
    \text{Prob}(a < \delta_2 c^*)\leq \text{Prob}(\mathbf{a}_{i, \mathcal{C}_k}^\top\mathbf{v}_{i, \mathcal{C}_k} \leq \delta_2 c^*) \leq  \epsilon_2,
\end{equation}
where $\mathbf{a}_{i, \mathcal{C}_k} = (\mathbf{\mu}_{\mathcal{C}_k} - \mathbf{x}_i) / \lVert \mathbf{\mu}_{\mathcal{C}_k} - \mathbf{x}_i \rVert$. It can be further transformed into
\begin{equation}
    \mathbf{a}_{i, \mathcal{C}_k}^\top (\mathbf{\mu}_{\mathcal{C}_k} - \mathbf{x}_i) - \delta_2 c^* \geq \text{erf}^{-1}(1 - 2\epsilon_2) \sqrt{2\mathbf{a}_{i, \mathcal{C}_k}^\top \mathbf{\Sigma}_{\mathcal{C}_k}\mathbf{a}_{i, \mathcal{C}_k}}.
    \label{eq:type_II_determin}
\end{equation}
We write this equation as $h_{\mathcal{C}_k}(\mathbf{x}_i) \geq 0$ for simplicity. 

\subsection{Mission Objective}
\label{sec:objective_update}
The objective function Eq.~\ref{eq:opt_objective} involves minimizing the uncertainty of target state estimation when robots are at resultant position $\mathbf{x}_{t+1}$. We let $f(\cdot, \cdot)$ be the \textit{trace} of the estimation covariance matrix for target positions calculated using $\mathbf{x}_{t+1}$ and $\hat{\mathbf{z}}_{t+1}$ with the latter being the estimated positions of targets at time $t+1$. 
In other words, the first term in Eq.~\ref{eq:opt_objective} tries to find robots' next-step positions that minimize the tracking error utilizing the team's predicted information of the targets. In the upcoming part, we show that the trace function is non-convex in optimization variables.


To estimate the targets' positions recursively, we assume each robot is equipped with one range and one bearing sensor, whose measurements are nonlinear in target positions. We use $\mathbf{z}_t$ to denote the positions of targets, with the position of a particular target $j$ being $\mathbf{z}_{j, t} = [x_{T_{j,t}},~y_{T_{j, t}}]^\top$,
and use $\mathbf{x}_{i,t} = [x_{i,t},
~y_{i,t}]^\top$ to denote the position of robot $i$. 
Based on the derivation in~\cite{zhou2011multirobot}, robot $i$'s range and bearing measurements of target $j$ are 
$d_{ij} = \sqrt{\Delta x_{ij}^2 + \Delta y_{ij}^2}$ and $\theta_{ij} = \text{arctan}(\frac{\Delta y_{ij}}{\Delta x_{ij}}) - \phi_i$,
where $\Delta x_{ij} = x_{T_j} - x_{i},{}\Delta y_{ij} = y_{T_j} - y_{i}$, and $\phi_i$ is the orientation of robot $i$. We drop the time step $t$ here for simplicity of notation. Robot $i$'s measurement of target $j$ is thus 
\begin{equation}
    \mathbf{y}_{ij} = \left[\begin{array}c 
    d_{ij} \\ \theta_{ij} 
    \end{array}\right] + \left[\begin{array}c 
    n_{d_{ij}} \\ n_{\theta_{ij}} 
    \end{array}\right].
\end{equation}
Here, $n_{d_{ij}}$ and $n_{\theta_{ij}}$ are noises in range and bearing measurements, respectively. We assume both follow zero-mean Gaussian distributions. Specifically, we let $n_{d_{ij}} \sim \mathcal{N}(0, R_{d, ij})$. The measurement noise covariance, $R_{d, ij}$, is a nonlinear function of the distance between robot $i$ and target $j$. We define it through its inverse as
    $R^{-1}_{d, ij} = a_d \cdot \text{exp}(-\lambda_d \lVert\mathbf{x}_{i} - \mathbf{z}_{j} \rVert)$,
where $a_d$ and $\lambda_d$ are constant parameters determining the characteristic of noise. The bearing measurement noise covariance, $R_{\theta, ij}$, is defined similarly. The team-level measurement noise covariance matrix with respect to target $j$, i.e., $\mathbf{R}_j$, is thereby constructed as $\mathbf{R}_j^{-1} = \text{diag}(R^{-1}_{d, 1j}, R^{-1}_{\theta, 1j}, R^{-1}_{d, 2j}, R^{-1}_{\theta, 2j}, \cdots, R^{-1}_{d, Mj}, R^{-1}_{\theta, Mj})$, 
and it follows that $\mathbf{R} = \text{diag}(R_1, R_2, \cdots, R_N)$ represents the noise covariance matrix of the whole robot team measuring all targets.

To compute the uncertainty of target position estimation, we need measurement matrix for a combination of range and bearing sensors. By~\cite{zhou2011multirobot}, we derive robot $i$'s measurement matrix of target $j$ as
\begin{equation}
    \mathbf{H}_{ij} = \left[\begin{array}c 
    \mathbf{h}_{d_{ij}}^\top \\ \mathbf{h}_{\theta_{ij}}^\top 
    \end{array}\right], 
\end{equation}
where $\mathbf{h}_{d_{ij}}$ and $\mathbf{h}_{\theta_{ij}}$ are column vectors corresponding to the range and bearing measurements. With $\mathbf{p}_{ij} = \mathbf{x}_i - \mathbf{z}_j$, the two measurement vectors are
\begin{equation}
    \mathbf{h}_{d_{ij}} = -\frac{\mathbf{p}_{ij}}{\sqrt{\mathbf{p}_{ij}^\top \mathbf{p}_{ij}}},
    \quad
    \mathbf{h}_{\theta_{ij}} = \frac{\mathbf{J}\mathbf{p}_{ij}}{\mathbf{p}_{ij}^\top \mathbf{p}_{ij}}.
\end{equation}
$\mathbf{J}$ is the $2 \times 2$ rotational matrix applied at $-\frac{\pi}{2}$. We can concatenate the measurement matrices of target $j$ from each robot vertically to obtain the whole team's measurement matrix as
$\mathbf{H}_j = [\mathbf{H}_{1j}; \mathbf{H}_{2j}; \cdots; \mathbf{H}_{Mj}]$.
The robot team's measurement matrix of all targets are thereby constructed as $\mathbf{H} = \text{diag}(\mathbf{H}_1, \mathbf{H}_2, \cdots, \mathbf{H}_N)$.

To compute the uncertainty in objective function Eq.~\ref{eq:opt_objective}, we perform one iteration of EKF to get the ``imagined'' uncertainty for target position estimation. First, in the \textit{prediction step} of EKF, the propagated estimate $\hat{\mathbf{z}}_{t+1|t}$ and the corresponding covariance matrix $ \mathbf{P}_{t+1|t}$ are, 
 \begin{equation}
    \hat{\mathbf{z}}_{t+1|t} = \mathbf{A}\hat{\mathbf{z}}_{t}, 
    \quad 
    \mathbf{P}_{t+1|t} = \mathbf{A}\mathbf{P}_{t}\mathbf{A}^\top + \mathbf{Q},
\end{equation}
where $\hat{\mathbf{z}}_t$ is the current estimate of target positions, $\mathbf{P}_t$ is the current estimation covariance matrix, and $\mathbf{A}$ is the process matrix of target motions. Then, the \textit{update step} of EKF further updates the covariance matrix using $\mathbf{R}$ and $\mathbf{H}$ we have derived above:
\begin{equation}
    \label{eq:cov_inv_update}
    \mathbf{P}^{-1}_{t+1|t+1} = \mathbf{P}^{-1}_{t+1|t} + \mathbf{H}^\top \mathbf{R}^{-1} \mathbf{H},
\end{equation}
With this, we can compute the trace of the covariance matrix $\text{Tr}(\mathbf{P}_{t+1|t+1})$ and apply it to evaluate the tracking performance. 

\subsection{Reformulated Optimization}

Putting all the pieces together, target tracking under the existence of danger zones is reformulated as the subsequent optimization:
\vspace{-0.1cm}
\begin{subequations}
    \label{eq:opt_approx}
    \begin{align}
       & \min_{\mathbf{u}_{t}} \ \   w_1\cdot \text{Tr}(\mathbf{P}_{t+1}) + w_2\cdot \sum_{i=1}^{M}\lVert \mathbf{u}_{i,t}\rVert \label{eq:opt_approx_objective}\\ 
         & \textrm{s.t.}~~~~\mathbf{x}_{t+1} = \mathbf{\Phi}\mathbf{x}_{t} + \mathbf{\Lambda}\mathbf{u}_{t}, \label{eq:opt_approx_dynamics}
         \\&~~~~~\quad
        g_{\mathcal{S}_l}(\mathbf{x}_{i, t+1}) \geq 0, \forall i \in [M], \forall l \in [p]
        \label{eq:opt_approx_constraint_1},\\
         &~~~~~\quad h_{\mathcal{C}_k}(\mathbf{x}_{i, t+1}) \geq 0, \forall i\in [M], \forall k\in [q].\label{eq:opt_approx_constraint_2}  
    \end{align}
\end{subequations}
We use a nonlinear solver Forces Pro~\cite{FORCESNLP, FORCESPro} to solve this nonlinear optimization and obtain control inputs for the robots at every time step.

\section{Simulations}
\vspace{-0.2cm}
\label{sec:simu_results}
 We evaluate the performance of our risk-aware target tracking framework under the setting where (i) there exists a sensing danger zone; (ii) there exists a communication danger zone, in MATLAB simulations\footnote{ We open source our code at:
\url{https://github.com/Zhourobotics/DZone_Tracking}}.

\vspace{-0.4cm}
\subsection{Sensing danger zones}

We conduct several simulations to show that robots display corresponding risk-aware behaviors in response to changes in uncertainty levels and risk requirements.
We provide numerical analysis in the scenario where 2 robots track 2 targets under the threat of one sensing danger zone in different conditions and uncertainty settings.  
The safety clearance is set as $r = 2$, and weights in objective Eq.~\ref{eq:opt_approx_objective} are set as $w_1 = 2.0, w_2 = 0.01$.

\begin{figure}[!ht]
    \centering
    \vspace{-0.3cm}
      \includegraphics[width=0.88\columnwidth]{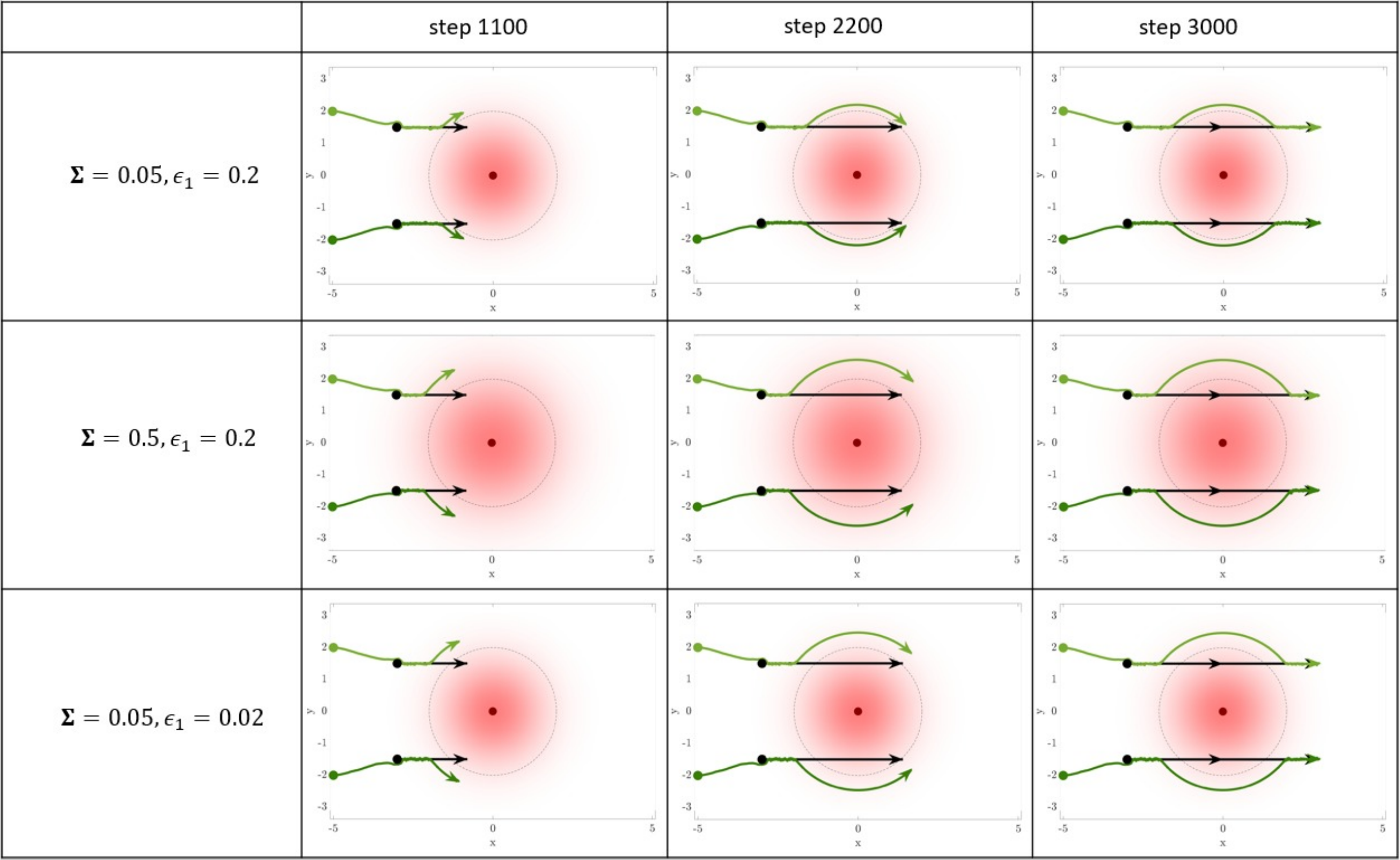}
    \caption{Risk-aware tracking with a sensing danger zone. Each row corresponds to one selection of the combination ($\mathbf{\Sigma}$, $\epsilon_1$), and three subfigures in the same row show the corresponding tracking process. Green lines are the trajectories of robots and black lines are the trajectories of targets.
    We use light and dark green for the two robots respectively. The initial state of robots and targets are drawn as dots with the corresponding colors. The red area represents a Gaussian distribution with an associated mean position in dark red dots. 
    A more spread-out distribution corresponds to a larger $\mathbf{\Sigma}$ in the second row. The dotted circle represents the safety clearance as the radius.}
    \label{fig:bm_typeI_qualitative}
    \vspace{-0.2cm}
\end{figure}

\vspace{-0.5cm}
The qualitative results containing trajectories of both robots and targets are shown in Fig.~\ref{fig:bm_typeI_qualitative}. 
Different condition parameters impact robots' trajectories when they are in proximity to the sensing danger zone. 
The subfigures in the first and second rows illustrate that the robots' behaviors become more conservative to keep a larger distance from the danger source with the increase in uncertainty $\mathbf{\Sigma}$.
Compared with the first parameter setting (first row), the third one decreases the value of $\epsilon_1$, i.e., it imposes stricter risk requirements on robots prohibiting them from entering the sensing danger zone. In this case, it shows that the robots also keep a larger distance away from the danger source.

\begin{figure}[!t]
    \centering
    \captionsetup[subfigure]{font=scriptsize, labelfont=scriptsize}
    \subfigure[Trace (\textbf{S})]{
      \includegraphics[width=0.22\columnwidth]{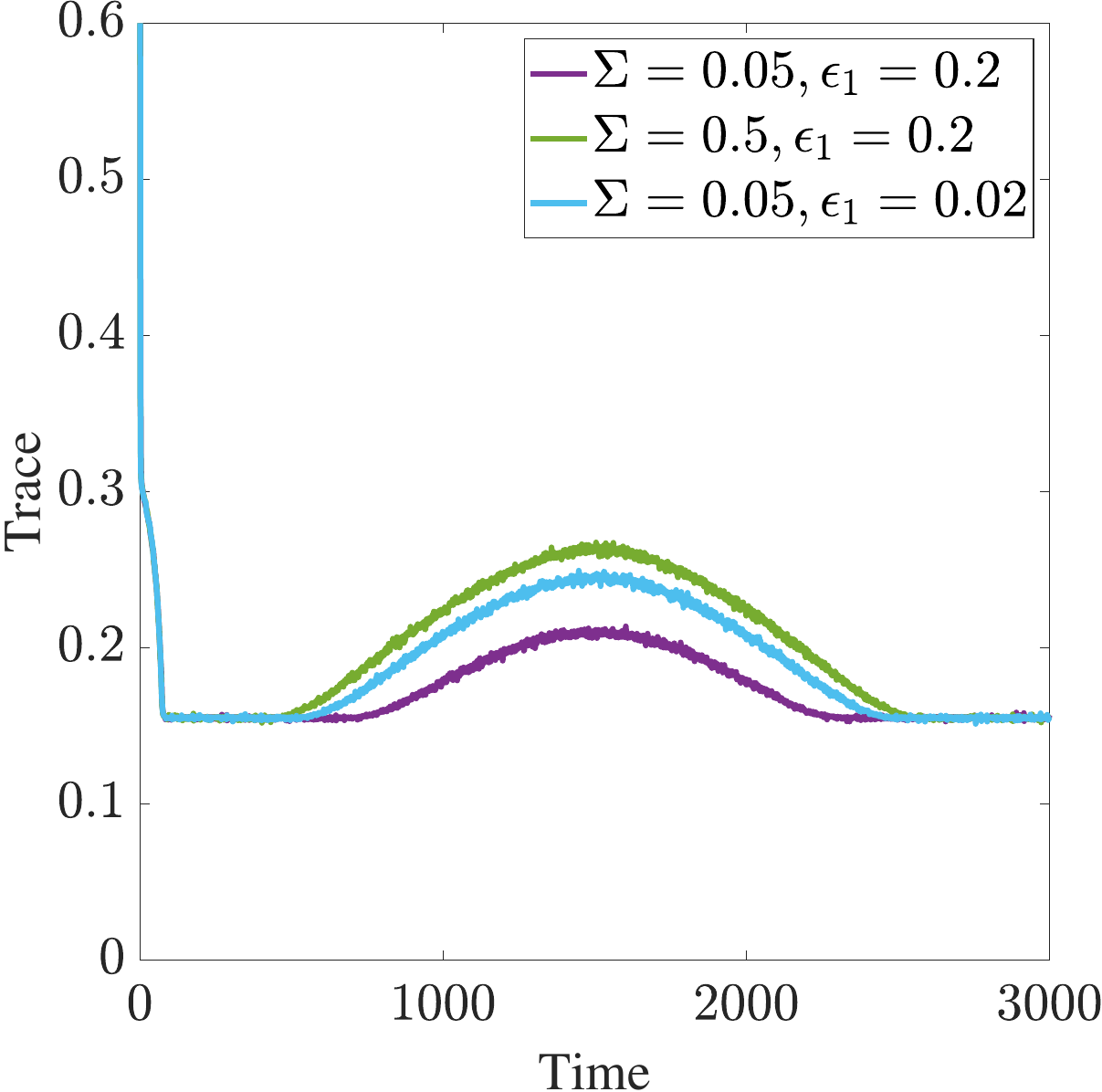}
      \label{fig:typeI_trace}
      \vspace{-0.4cm}
    }
    \subfigure[Probability (\textbf{S})]{
      \includegraphics[width=0.22\columnwidth]{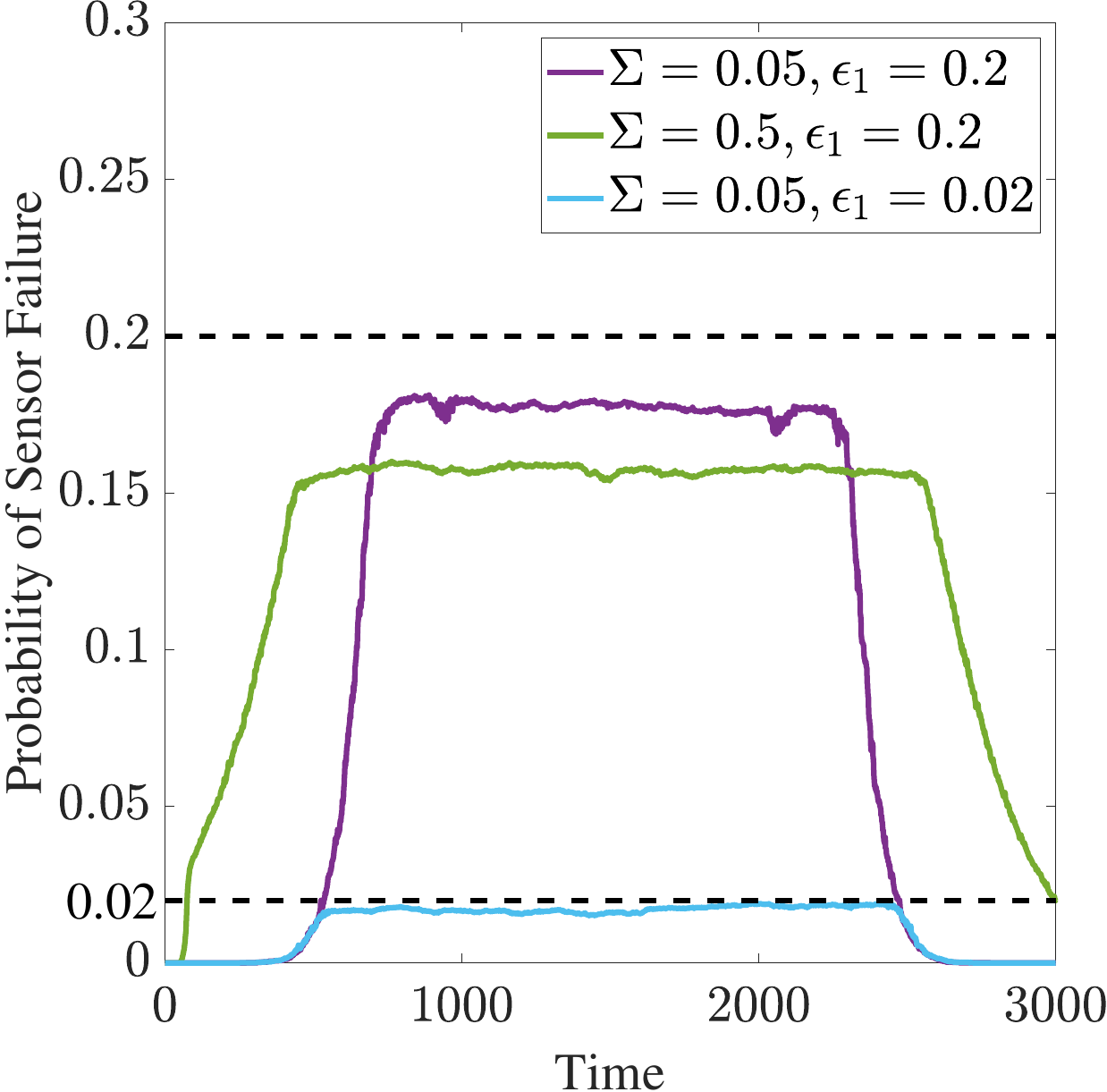}
      \label{fig:typeI_risk}
    }
    \subfigure[Trace (\textbf{C})]{
      \includegraphics[width=0.21\columnwidth]{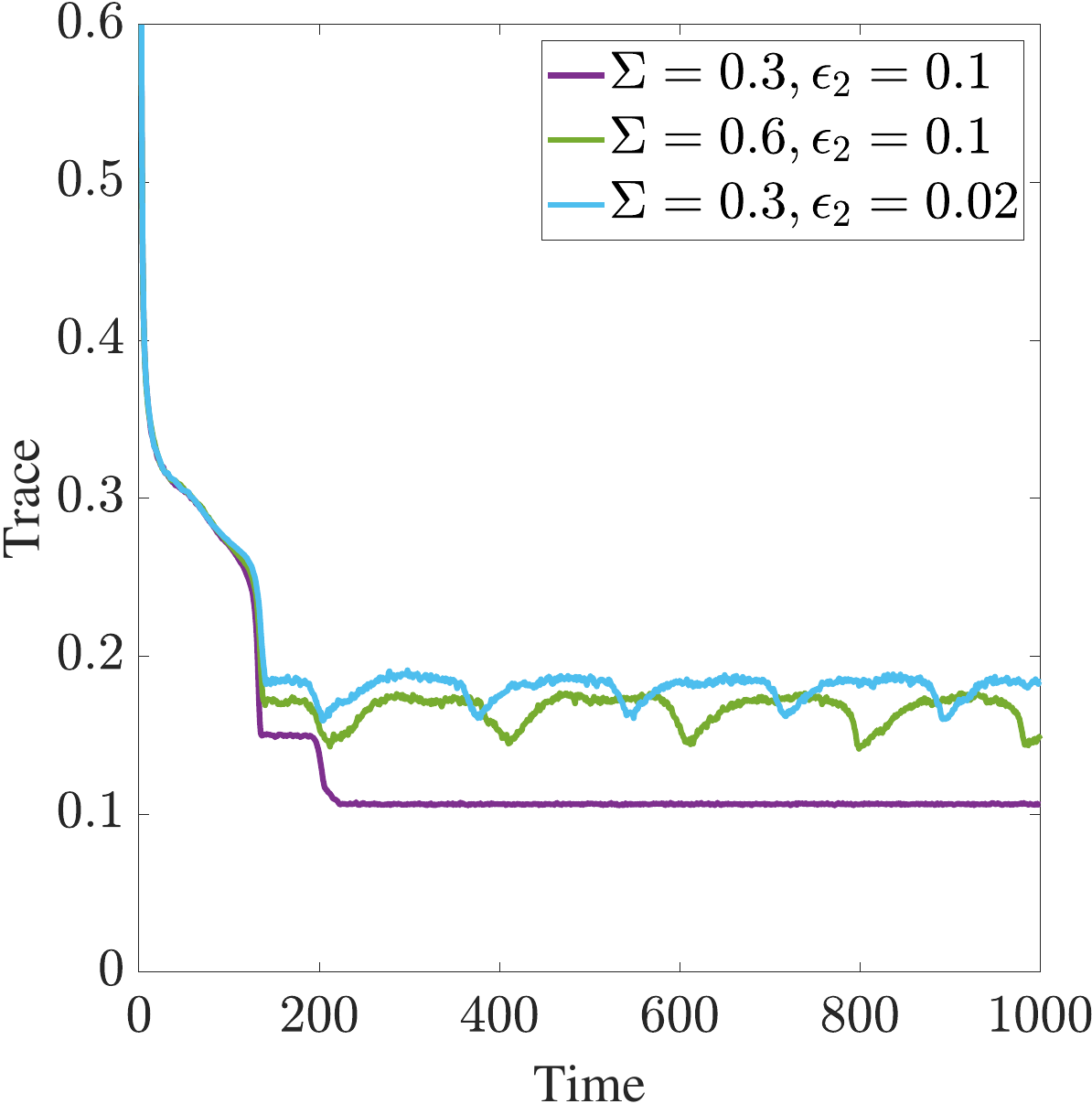}
      \label{fig:typeII_trace}
      \vspace{-0.4cm}
    }
    \subfigure[Probability (\textbf{C})]{
      \includegraphics[width=0.22\columnwidth]{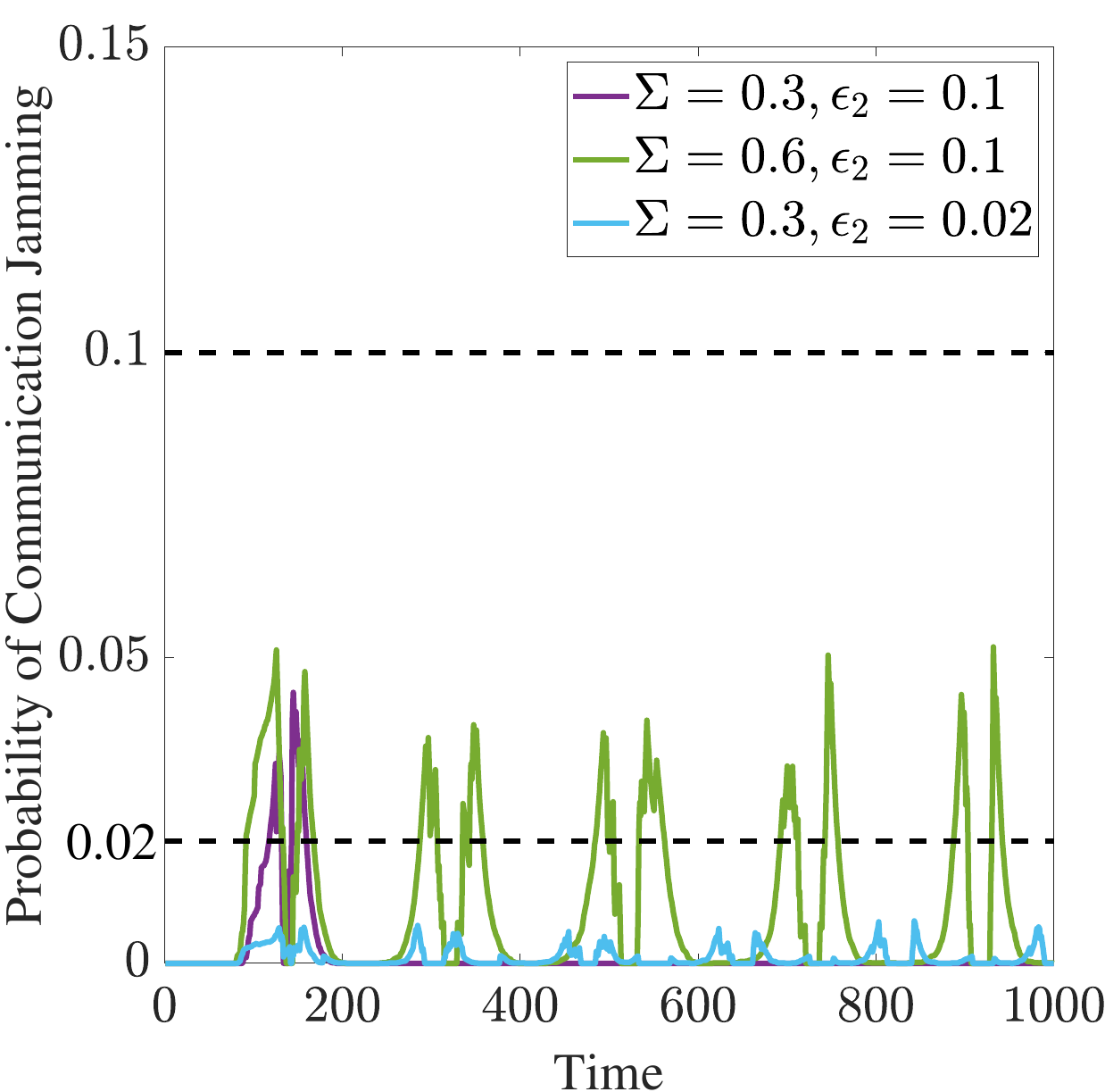}
      \label{fig:typeII_risk}
    }
    \caption{(a)(b): trace and probability of sensor failure throughout the tracking process with a \textbf{S}ensing danger zone; (c)(d) trace and probability of communication jamming with a \textbf{C}ommunication danger zone. The comparison of results under three sets of parameters is shown.}
    \label{fig:quant}
    \vspace{-0.4cm}
\end{figure}

We further present the quantitative results of three comparative simulations in Fig.~\ref{fig:quant}. 
Trace value encodes tracking performance, while the probability of failure encodes how dangerous it is for the robots. 
In Fig.~\ref{fig:typeI_trace}, the tracking performance deteriorates when the robots take a detour to circumvent the danger zone but resumes as robots leave the zone. 
As the robots take a detour, the scenario with $\mathbf{\Sigma} = 0.05, \epsilon_1 = 0.2$ sustains a smaller trace and better tracking performance than the other two, which is consistent with the qualitative results.

Fig.~\ref{fig:typeI_risk} shows the probability of sensor failure, defined as the probability that robots are within the safety clearance of the danger source. 
Since it is not straightforward to compute the probability, we estimate it through sampling. 
We first sample the actual position of the danger source for $1000$ times from its distribution. 
Using each sample, we check whether the distance between a robot and the danger source is less than the corresponding safety clearance. If so, the robot is considered to have a sensor failure.
Suppose, out of 1000 samples, a robot is within the zone for $\beta$ times, then the approximated probability is $\frac{\beta}{1000}$. We then average the probabilities across all robots. Results in Fig.~\ref{fig:typeI_risk} show that risk requirement (i.e., the probability lower than $\epsilon_1$) is met throughout the whole process for all three settings. The probability of sensor failure is lowest in the third case, which is reasonable because it has the smallest $\epsilon_1$ value and is the strictest in enforcing the risk requirement. The first case has the largest overall probability of failure, but it is still obviously smaller than $\epsilon_1 = 0.2$. 
\subsection{Communication danger zones}
We also evaluate the tracking performance under one communication danger zone to illustrate how an uncertain communication danger zone influences robots' risk-aware behavior. 
In a setting where 4 robots track 2 targets, we adopt three different combinations of parameters including the uncertainty of the jamming source's position $\mathbf{\Sigma}$ and risk requirement $\epsilon_2$ to compare robot trajectories. The two targets follow a circular motion inside a communication danger zone. 
The weights in the objective function (refer to Eq.~\ref{eq:opt_objective}) are $w_1 = 2.0, w_2 = 0.002$.

\begin{figure}[!ht]
\vspace{-0.4cm}
    \centering
      \includegraphics[width=0.88\columnwidth]{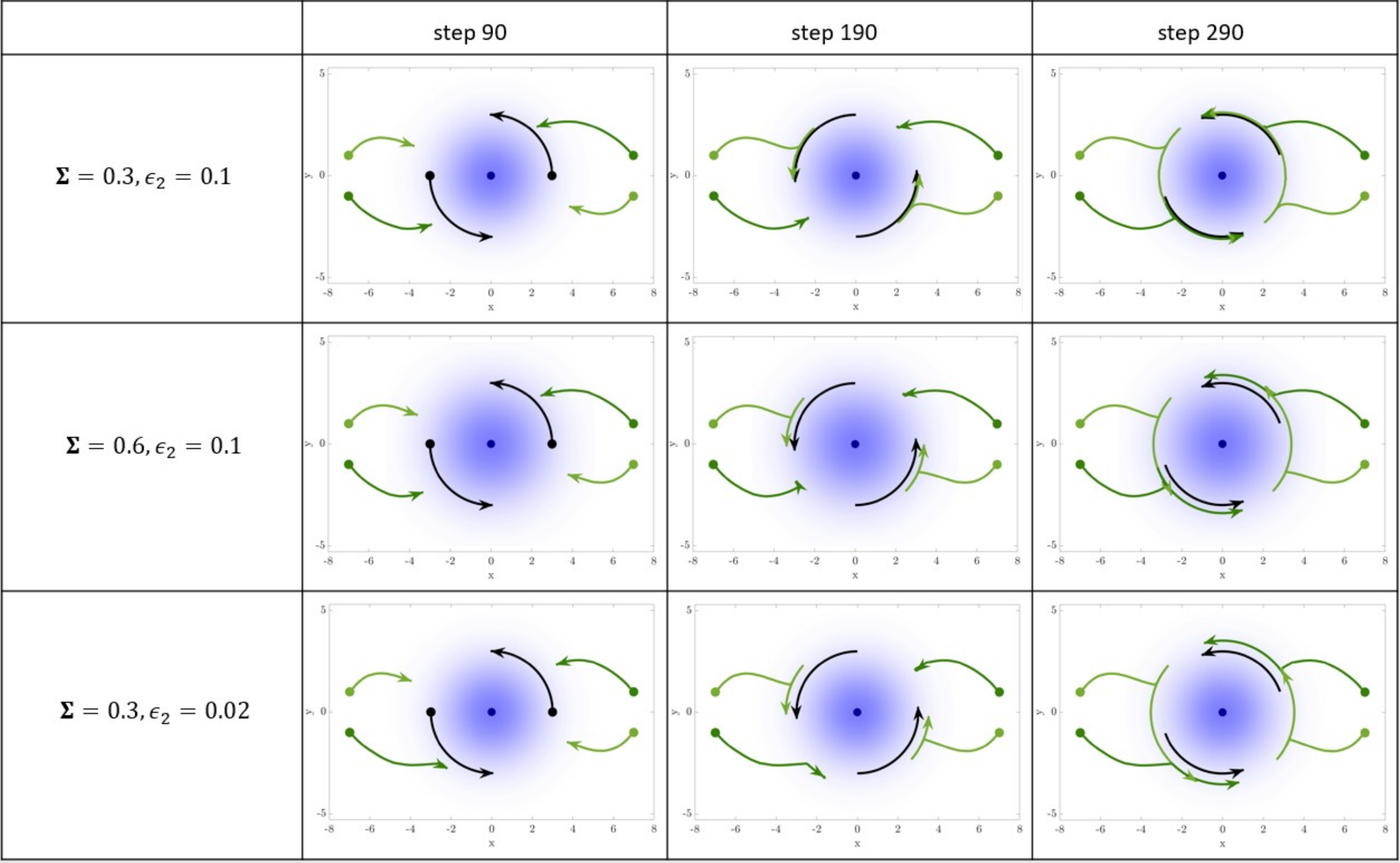}
    \caption{Risk-aware tracking with a communication danger zone. Each row corresponds to a tracking case under one parameter setting, and three sub-figures in the same row show the tracking process under that setting. Robots' trajectories are plotted in dark and light green, and the targets' trajectories are plotted in black. The two targets move in circles in a counter-clockwise direction. Robots' initial positions are shown as green dots. The first column also shows the initial positions of targets as black dots. The blue area represents that the position of the jamming source follows a Gaussian distribution. Uncertainty of the jamming source's position is reflected by how spread-out the area is, i.e., a larger $\mathbf{\Sigma}$ leads to a more spread-out area. Dark blue dots denote the mean position of the jamming source. }
    \label{fig:bm_typeII_qualitative}
    \vspace{-0.4cm}
\end{figure}

We visualize the trajectories of robots and targets in Fig.~\ref{fig:bm_typeII_qualitative}.
Under all three parameter settings, the robots can successfully track and follow the targets. The four robots divide themselves into two subgroups, with one subgroup following one target. The two robots in the same subgroup stay close to each other to secure their communications from jamming. Comparing the first and second cases shows that increasing the uncertainty in the jamming source's position forces the robots to stay further away from the communication danger zone. This aligns with our intuition since if the jamming source's position is more uncertain, it is safer for robots to keep a larger distance from the jamming source to protect and maintain inter-robot communications. A comparison of the third case and the first case shows that decreasing the required risk level $\epsilon_2$ has a similar effect, as the robots need to maintain a larger distance from the jamming source to meet the stricter risk requirement.

The corresponding quantitative results, consisting of trace and the probability of communication jamming, are presented in Fig.~\ref{fig:quant}. 
Fig.~\ref{fig:typeII_trace} verifies that either increasing the uncertainty level $\mathbf{\Sigma}$ or decreasing the required risk level $\epsilon_2$ steer robots away from the communication danger zone. This prevents them from closely following the targets, and consequently, decreases the tracking performance. 
The probability of jamming as shown in Fig.~\ref{fig:typeII_risk} is estimated through sampling. Specifically, we sample the position of the jamming source 1000 times from the Gaussian distribution it follows. For a particular robot $i$, we calculate its distance to the sampled position, $a$, and its distance from its furthest teammate $c^*$ (see Eq.~\ref{eq:max_dist}). Particularly, we use only teammates that are within a communication range, i.e., neighbors, to calculate this distance. Then, for each sample, we test whether $a < \delta_2 c^*$ holds. If the condition holds, we consider robot $i$ to be jammed. Otherwise, we consider it to be safe. Supposing the condition check holds for $\beta$ times, then the estimated probability is $\frac{\beta}{1000}$. We average the probability value across all robots to obtain the results plotted in Fig.~\ref{fig:typeII_risk}. As seen, the approximated probability of communication jamming remains below the required upper-bound $\epsilon_2$ (shown in dotted lines) for all three cases. Notably, for the third case, the approximated probability remains nearly $0$ throughout the whole tracking process since the risk requirement $\epsilon_2 = 0.02$ is stringent.


\vspace{-0.3cm}
\section{Experiments}
\vspace{-0.2cm}
We conducted several hardware experiments to evaluate the performance under real-world tracking uncertainty and dangerous environments.
We use a team of Crazyflie drones~\cite{crazyflie} as trackers and multiple Yahboom ROSMASTER X3 ground robots~\cite{yahboom} as targets.
Danger zones are set up as paper disks on the floor of an indoor environment. 
We consider the ground targets with non-adversarial movements. The target trajectories are prespecified but trackers are agnostic to the movement pattern.
Trackers need to estimate the positions of ground targets and follow them closely while satisfying the risk requirements.
The optimal control inputs are solved online in Eq.~\ref{eq:opt_approx} at every time step on a local Desktop and sent to Crazyflies. Eq.~\ref{eq:opt_approx} is solved within 50ms on average. Since we are not imposing inter-robot collision avoidance constraints, we make the trackers fly at different fixed heights.

\begin{figure}[!ht]
    \centering
    \subfigure[]{
      \includegraphics[height=3.2cm]{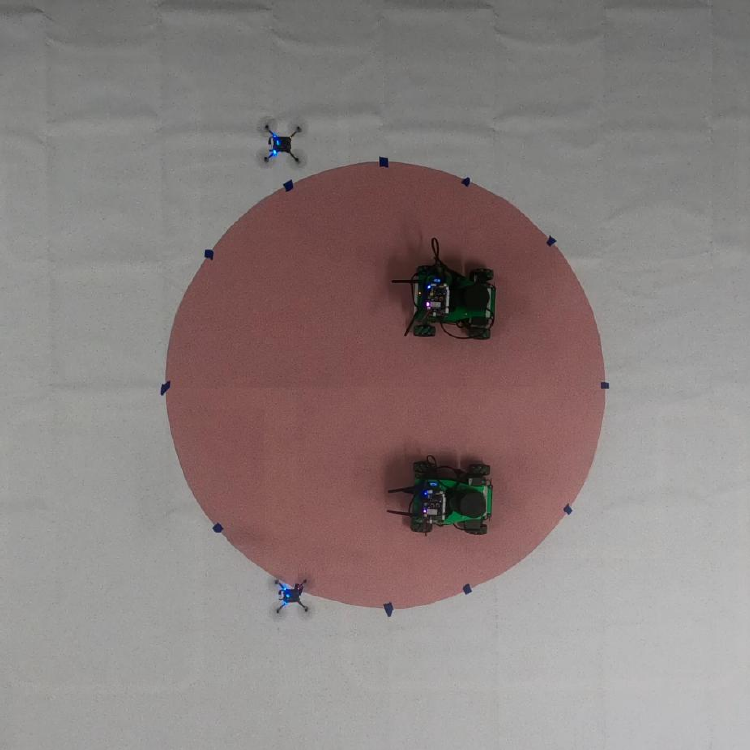}
      \vspace{-0.4cm}
    }
    \subfigure[]{
      \includegraphics[height=3.2cm]{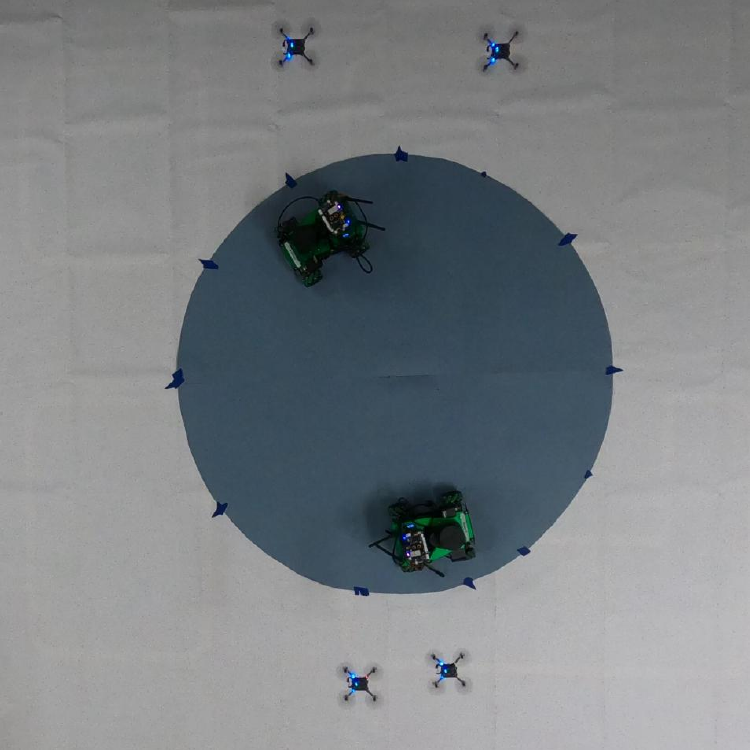}
      \vspace{-0.4cm}
    }
    \subfigure[]{
      \includegraphics[height=3.2cm]{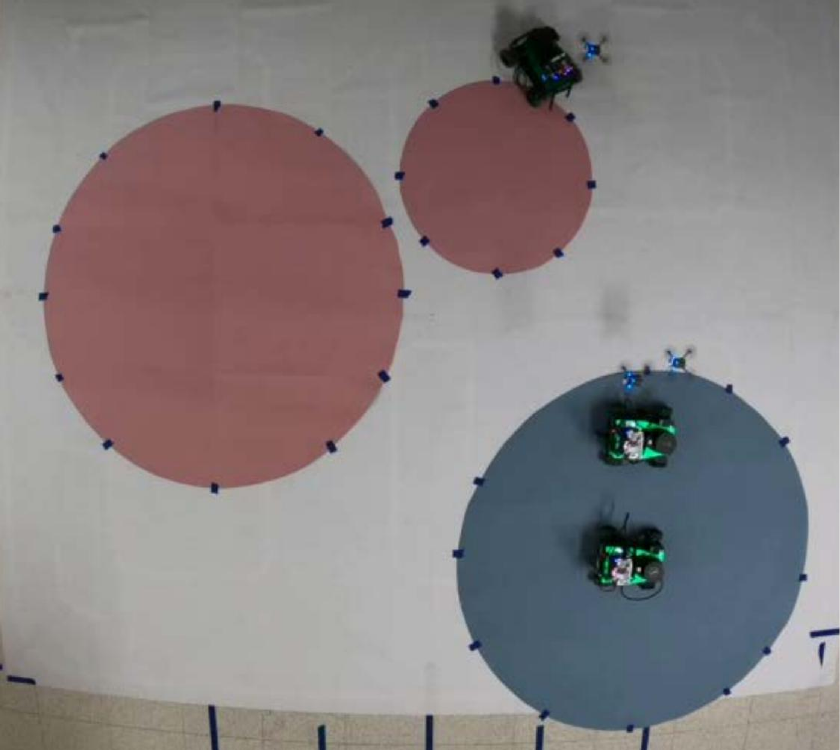}
      \vspace{-0.4cm}
    }
    \caption{Screenshots of hardware experiments. (a) 2 robots track 2 targets with a sensing danger zone; (b) 4 robots track 2 targets with a communication danger zone; (c) 3 robots track 3 targets with both types of danger zones. The video of the experiments is available at: \url{https://youtu.be/uSYPI817Y6c}} 
    \label{fig:hardware_scene}
    \vspace{-0.7cm}
\end{figure}

We validated the simulation scenarios from Sec.~\ref{sec:simu_results} on the real-robot system and showed the risk-aware tracking behaviors in Fig.~\ref{fig:hardware_scene}. The parameters including $\epsilon_1$, $\epsilon_2$, $\mathbf{\Sigma}$, and the safety clearance of the sensing danger zones are adjusted to suit the indoor experimental conditions. Fig.~\ref{fig:hardware_scene}(a) and Fig.~\ref{fig:hardware_scene}(b) display the robots' behaviors around a sensing and a communication danger zone, respectively. 
The robots demonstrate risk awareness in their actions. 
When the targets are inside the sensing danger zone, robots take a detour and keep a distance to ensure their safety. 
Similarly, when targets approach the communication danger zone, pairs of robots tracking the same target stay close to each other to secure their communication links.

To demonstrate the robustness of our proposed method, we perform an experiment where three robots track three targets, with two sensing and one communication danger zones. 
As shown in Fig.~\ref{fig:hardware_scene}(c), one target at the top moves in a circular motion while the two targets at the bottom move with constant velocities from left to right. 
The robots validate risk awareness through their behaviors: the top robot avoids the sensing danger zone even when the target is inside it, while the two bottom robots stick together closely to prevent communication jamming near the communication danger zone.

During the experiments, due to imprecise localization and actuation uncertainty, we occasionally run into situations where trackers end up being within a danger zone at some time steps and the reformulated optimization in Eq.~\ref{eq:opt_approx} become infeasible. This demonstrates the difficulty of sim-to-real transfer for multi-robot target tracking. To address this situation, we apply a single-step control input such that the trackers can resiliently escape from danger zones. The escape control points towards the direction from the mean position of the danger source to tracker's current position. After escaping from the zone, trackers resume their regular planning procedure. 

\vspace{-0.1cm}
\section{Conclusion and Discussion}
\vspace{-0.2cm}


In this paper, we proposed a risk-aware multi-robot target tracking framework in danger zones.
We modeled the danger zones into different attack types and formulated the optimal control as a nonlinear optimization problem. 
The problem is reformulated with approximated chance constraints and tracking mission objectives. 
We verified in both simulations and hardware experiments that with different uncertainty levels and risk requirements, robots exhibit corresponding risk-aware behaviors in proximity to danger zones. 
For real-world deployments with the condition that robots fall into the danger zones, we provide a practical method to escape from the danger zone to continue tracking task. 
In the future, we will extend the work for resilient tracking and address the perception and joint localization in target tracking~\cite{10155978} for better sim-to-real transfer. 

\bibliography{author.bib}

\end{document}